\documentclass{article}

    \PassOptionsToPackage{numbers, compress}{natbib}
\usepackage[preprint]{neurips_2020}

\usepackage{hyperref}
\hypersetup{
    colorlinks=true,
    linkcolor=blue,
    filecolor=magenta,      
    urlcolor=blue,
}
\usepackage[utf8]{inputenc} % allow utf-8 input
\usepackage[T1]{fontenc}    % use 8-bit T1 fonts
\usepackage{hyperref}       % hyperlinks
\usepackage{url}            % simple URL typesetting
\usepackage{booktabs}       % professional-quality tables
\usepackage{amsfonts}       % blackboard math symbols
\usepackage{nicefrac}       % compact symbols for 1/2, etc.
\usepackage{microtype}      % microtypography

\usepackage{caption}
\usepackage{graphicx}
\usepackage{enumitem}

\title{Audeo: Audio Generation for a Silent Performance Video}
\author{
  Kun Su\\
  Department of Electrical and Computer Engineering\\
  University of Washington, Seattle, WA 98195\\
  \AND
  Xiulong Liu \\
  Department of Electrical and Computer Engineering\\
  University of Washington, Seattle, WA 98195\\
  \AND
  Eli Shlizerman \\ 
  Department of Applied Mathematics \\
  Department of Electrical and Computer Engineering\\ 
  University of Washington, Seattle, WA 98195
}

\begin{document}

\maketitle
% \begin{center}
%     \includegraphics[width=\linewidth]{figure_1.png}
%     \captionof{figure}{Audeo: A full pipeline of audio generation for a silent performance video.}
% \end{center}
\begin{abstract}
We present a novel system that gets as an input video frames of a musician playing the piano and generates the music for that video. Generation of music from visual cues is a challenging problem and it is not clear whether it is an attainable goal at all. Our main aim in this work is to explore the plausibility of such a transformation and to identify cues and components able to carry the association of sounds with visual events. To achieve the transformation we built a full pipeline named `\textit{Audeo}' containing three components. We first translate the video frames of the keyboard and the musician hand movements into raw mechanical musical symbolic representation Piano-Roll (Roll) for each video frame which represents the keys pressed at each time step. We then adapt the Roll to be amenable for audio synthesis by including temporal correlations. This step turns out to be critical for meaningful audio generation. As a last step, we implement Midi synthesizers to generate realistic music. \textit{Audeo} converts video to audio smoothly and clearly with only a few setup constraints. We evaluate \textit{Audeo} on `in the wild' piano performance videos and obtain that their generated music is of reasonable audio quality and can be successfully recognized with high precision by popular music identification software. 
% An audio-visual event requires vision and sound channels to be fully described. Missing audio will lead to incorrect understanding of the actual event especially in music performance video. In this paper, we investigate whether we can generate music from a silent piano performance video only. We introduce a full pipeline named "Audeo" containing three stages to achieve this challenging goal. Given visual information only, we first translate the piano performance to raw musical symbolic representation (MIDI) at each video frame. The initial MIDI is then improved by including temporal correlation as together. In the end, a MIDI synthesizer network will be used to generate realistic musical audio. Audeo converts video to audio smoothly and clearly with few setup constraint. We evaluate Audeo using piano performance video in the wild and generated audios can successfully deceive music identification software.
\end{abstract}

\section{Introduction}
\begin{flushright}
\textit{Melody is the essence of music. I compare a good melodist to a fine racer.\\ Wolfagang Amadeus Mozart
}\end{flushright}

The perfect combination of musician's skills with the instrument's sounds creates the delightful experience of `live music'. Such an event is inspiring from the perspective of the melody being played and also from the perspective of witnessing an admirable synchrony between the musician and the instrument. 
% `I have never heard someone playing the piano so beautifully'. 
What makes the musical performance to sound as it sounds? The answer to this question is interwined. We know many of the ingredients that make musical performance to sound well, however, we do not know how to rigorously quantify the contribution of the components. Notes, tempo, consistency, timed precision, mechanical accurateness, rhythmic movements, harmonics, frequencies; all these and many more delicately compose the melody of a musical piece. Quantifying these aspects plays a key role in an attempt to better understand how a realistic melody is generated.

% Indeed, vision and sound have strong correlation and in many applications and both of them contribute to the perception of the event. An evidence for their importance is that when one of them is perturbed this significantly impairs the understanding of the actual audio-visual event, e.g. watching a speech on television with asynchronous audio and video impairs our ability to associate lips movement and the words being said affecting our perception of the content.

A particular test to inform us regarding music generation is the attempt to constitute the music from visual information, i.e., finding possible ways to recreate the audio stream of a musical performance, just from the visual stream. In the case of a piano recording, that would be taking into account the positions of the musician's hands and the body and the positions of the keys and the pedals and merge them together into music. In such an endeavor, timed precision between visual cues and sounds is known to have a profound effect and takes the form of more complex process than a mere synchronization. The reasons for the complexity stem from visual stream perception being of significantly slower rate than the perception of an audio stream, however, the perception of their combination requires a latency between audio and video signals to be faster than the rate of the visual stream. This creates an effect in which for generation of an audio signal for a video, one should not only find association between video frames and audio but also to precisely complete the audio stream in between the video frames going back an forth into the past and the future frames. Such completion is nontrivial and requires whether exhaustive knowledge of the instrument and body mechanics, i.e., a virtual instrument, or an ability to imagine the details from visual features, similar to composer's ability to envision the melody from reading musical notes.

Frames of a video include an abundance of visual information, some of which could be irrelevant to music. It is therefore plausible that instead of a direct transformation, intermediate features could be used for the translation from video to audio. These features should capture the mechanical and the perceptual features of the interaction between the musician and the instrument and be constructive tools for sound representation and synthesis. For example, Musical Instrument Digital Interface (Midi) protocol is a candidate signal. It is used for the interchange of musical information between instruments and encodes various keyboard functions and musical attributes. Variants of Midi, such as Pseudo-Midi will provide even more compact version to encode keyboard function and musical attributes altogether. Moreover, connecting visual actions with frequencies of the audio signal as it varies with time, i.e., the Spectrogram, can be a useful mediator.

\begin{figure}[t]
    \centering
    \includegraphics[width=\linewidth]{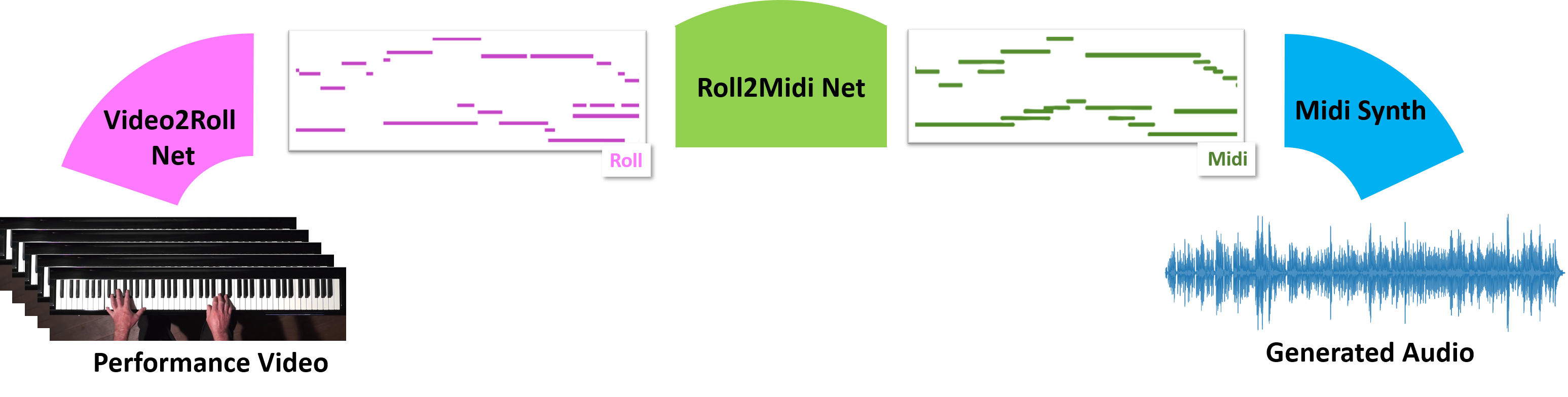}
    \caption{Given an input of video frames of musician playing the piano, \textit{Audeo} generates the music for that video. Please also see supplementary video and materials with sample results. Keyboard images are from \href{https://www.youtube.com/user/PaulBartonPiano}{Paul Barton YouTube Channel} with permission.}
    \label{fig:figure_1}
\end{figure}

In this work we address the challenge of music generation from video by proposing a full pipeline, \textit{Audeo}, to generate the audio of a silent piano performance video. \textit{Audeo} translates the performance from video to audio domain in three stages through recovery of mediator signals. In the first stage, given a top-view video, we use a multi-scale feature attention deep residual network to capture visual information and predict which keys are pressed at each frame (Video2Roll Net). We formulate this as a multi-label classification task and the collection of predictions can be seen as a `Piano-Roll' \cite{muller2015fundamentals}. In the second stage, we utilize a Generative Adversarial Network (GAN)\cite{goodfellow2014generative} to refine and enhance the Roll with musical attributes which outputs the Pseudo-Midi signal (Roll2Midi Net). This step turns out to be critical for providing symbolic musical representation rather than mechanistic keyboard representation. The third and last stage of \textit{Audeo} pipeline is the synthesization of Pseudo-Midi to audio signal (Midi Synth). This step synthesizes the audio using a classical Midi synthesizer or using a deep synthesizer for more realistic output. The deep synthesizer translates Midi to a spectrogram and then to audio. An overview of our \textit{Audeo} system is shown in Fig. \ref{fig:figure_1}. Our main contributions are the following:
%\setlist{nolistsep}
%\begin{enumerate}[label=(\roman*)]
(i) To the best of our knowledge, we are the first work to generate music audio from `in the wild' silent piano performance videos.
(ii) We introduce a full pipeline named \textit{Audeo} containing three interpretable components to complete this transformation.
(iii) \textit{Audeo} is robust and generalizable. The generated audio of `in the wild' piano performance videos are detected well with popular music identification software.
%\end{enumerate}

\section{Related Work}
% \subsection{General Audio-Visual Cross-Domain Tasks}
While audio-visual signals are interrelated, classically there has been a clear separation of these signals into a single domain of video or audio. Deep learning approaches have succeeded to connect the two streams and to begin consideration of audio-visual tasks. Systems have been proposed to leverage and explore the correlation of both audio and video at the same time, i.e., \textit{audio-visual cross-domain tasks}. For example, conditioning the visual and sound streams on each other as a supervision in training was shown as an effective training method for networks with unlabeled data in audio-visual correspondence~ \cite{arandjelovic2017look,aytar2016soundnet,harwath2016unsupervised,owens2016ambient}. Moreover, it was shown that it is possible to separate object sounds by inspecting the video stream of an unlabeled video~\cite{gao2018learning} or to perform audio-visual event localization task on unconstrained videos~\cite{tian2018audio} and even to generate natural sounds, e.g., baby crying , water flowing, given a visual scene~\cite{zhou2018visual}. Such a generation task is conceptually similar to the task that we consider, however, it is on a much slower scale than generation of music and the visual input in the case of the piano playing has more structural content which needs to be processed.

Each direction of audio and video relation has been studied as well. In the \textit{audio-to-video} direction, deep learning RNN based strategies were proposed to generate body dynamics correlated with sounds from audio only ~\cite{shlizerman2018audio, lee2019dancing,ginosar2019learning}. Moreover, systems that generate parts of the face or synchronize lips movements from speech audio where shown to be possible~\cite{suwajanakorn2017synthesizing,oh2019speech2face}. In the \textit{video-to-audio} direction, prior work addressed the identification of objects most correlated with sounds. For piano performance that would be the keyboard, musician's hands, etc. Combination of traditional computer vision techniques were presented to provide these functionalities~\cite{akbari2015real,suteparuk2014detection,RobertMcCaffrey}. However, these methods turned out to be sensitive to the environment setup such as the camera position, illumination condition and so on. To improve performance, the use of depth cameras was proposed to detect the pressed keys with  depth information, however, while it indeed improved accuracy, such strategy cannot be generalized to unconstrained videos~\cite{nisbetcapture,rho2014automatic}. Furthermore, machine learning method such as Support Vector Machine (SVM)~\cite{suykens1999least} were proposed to classify a single key status, whether it is pressed or not~\cite{deb2016image}. Since these methods required a large manually labeled data to be trained on, systems using deep learning methods, such as Convolution Neural Networks (CNN), have been applied to key identification problem as well, approaching the problem as a binary classification task where each single key needs to be cropped separately and manually labeled before training and testing~\cite{akbari2018real,kang2019virtual}. Deep learning strategies also addressed both audio and video streams to identify actions such as following musical notes, i.e., a two-stream CNN has been proposed to determine the notes being played at any moment for the task of identifying whether correct fingers are pressed for the corresponding notes~\cite{lee2019observing}. 

The methods described above require a training set with an associated \textit{Ground truth Midi}. Such Midi is typically obtained with an electronic keyboard, a process that make creation of the training data to be limited. To overcome this challenge, the Onsets and Frames framework enables to transcript audio waveform to Midi~\cite{hawthorne2017onsets}. A recent work used this framework to obtain Pseudo Ground truth Midi and implemented a ResNet~\cite{he2016deep}, to predict the pitch onsets events (times and identities of keys  being pressed) given video frames stream~\cite{koepke2020sight}. While this method achieves acceptable onsets prediction there still exist a gap between the onset prediction problem and reconstruction of a complete Midi containing the offset information and from which music can be generated. \textit{Audeo} is using the Onsets and Frames framework to obtain a Pseudo Ground truth Midi for training as well and thus can be applied to any top-view video. Moreover, \textit{Audeo} generalizes the prediction task and generates a complete and robust Midi for synthesis via either traditional or deep learning based Midi synthesizers.

% Eli HERE
%\subsection{Music Generation}
% In synthesizing music from Midi, several deep learning approaches have been introduced.
In music generation, several deep learning approaches have been introduced. Autoregressive models which directly work on audio waveform such as Wavenet \cite{oord2016wavenet}, SampleRNN \cite{mehri2016samplernn} and their variants \cite{oord2017parallel,ping2018clarinet} have shown successes in both speech and music generation. However, transformation between two different domains (e.g. text to speech (TTS), symbolic musical score to audio) is more challenging. Tacotron \cite{wang2017tacotron,shen2018natural} proposed the encoder-decoder architecture to translate text to mel-spectrogram and a Wavenet conditioned on generated mel-spectrogram to generate final human speech waveform. In addition, Timbretron \cite{huang2018timbretron} uses CycleGAN \cite{zhu2017unpaired} for timbre style transform on spectrogram level. Recently, non-autoregressive models like MelGAN \cite{kumar2019melgan} also demonstrated convincing results on audio generation. However, unlike common human speech which is monophonic, piano music is challenging to generate due to its polyphonic property. In addition, symbolic Midi can be viewed as a time-frequency representation (while the text transcript for speech cannot). Since music is polyphonic and contains more content information, the TTS model cannot be directly applied to score-to-audio generation. While a conditional Wavenet has been proposed to enable Midi synthesis \cite{hawthorne2018enabling}, training a conditional Wavenet requires exhaustive computation resources. Another efficient possibility is to use Performance-Net (PerfNet) \cite{wang2019performancenet} which has been shown successfully converting Midi to spectrogram with computation efficiency. The last step of \textit{Audeo} uses the a pretrained PerfNet as deep learning based Midi synthesizer to generate audio in the spectrogram domain.

\section{Methods}
Our key approach is to use generalizable and interpretable mediator signals to translate piano video frames to output audio. Indeed, for piano performance videos from Internet (usually without accompanied ground truth Midi). We retrieve the Pseudo ground truth (GT) Midi from audio with the Onset and Frames framework \cite{hawthorne2017onsets}. This allows us to avoid hardware constraints of the instrument and to use any video, even those recorded in an unconstrained setup. The Pseudo GT Midi can be considered as a two dimensional binary matrix $M\in \mathbb{R}^{K\times T}$ where $K$ is the number of pitches and $T$ is the number of frames. For each entry $M_{k,j}$, $1$ indicates if the key $k$ is sustained at frame $j$ and $0$ otherwise. We describe the details of each component of \textit{Audeo} system in the following subsections.

\textbf{Video2Roll Net:}
\begin{figure}
    \centering
    \includegraphics[width=0.6\linewidth]{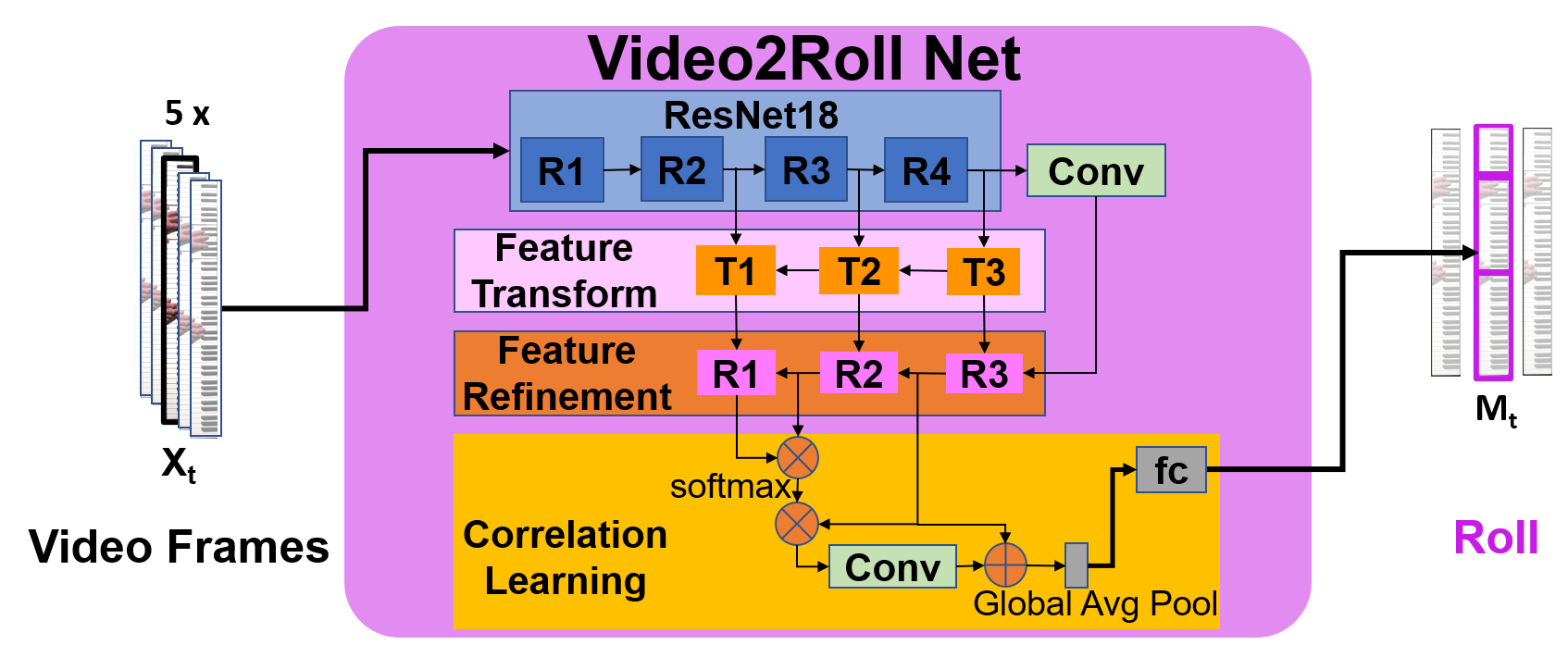}
    \caption{Detailed schematics of the components in VIDEO2Roll Net: ResNet18 + feature transform, feature refinement and correlation learning. Input: $5$ consecutive frames; Output: pressed-key prediction at the middle frame.}
    \label{fig:VIDEO2ROLL}
\end{figure}
\begin{figure}
    \centering
    \includegraphics[width=0.9\linewidth]{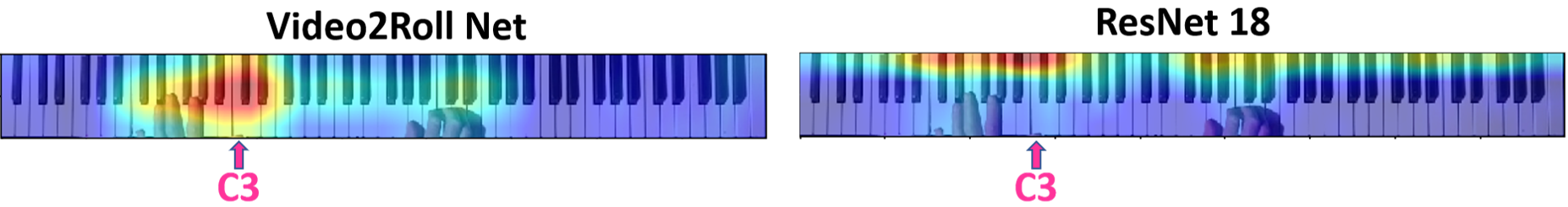}
    \caption{Visualized feature maps comparison between Video2Roll Net (left) and ResNet18 (right) using Scored Weighted Class Activation Heatmap (Score-CAM) \cite{wang2019score}. This example demonstrates that our method can locate the delicate visual cues of pressed C3 key more accurately. Keyboard images are from \href{https://www.youtube.com/user/PaulBartonPiano}{Paul Barton YouTube Channel} with permission.}
    \label{fig:feature_map}
\end{figure}
The task in this stage can be defined as a multi-label image classification problem. One video clip $X$ can be seen as a four dimensional matrix and each frame is $X_{t,c,h,w} \in \mathbb{R}^{T\times C \times H \times W}$ where $T$, $C$, $H$, $W$ are time, channel, height and width dimension respectively. We use stacked five consecutive gray scale frames $X_{t-2,t-1,t,t+1,t+2}$ as input into Video2Roll Net which outputs a prediction of the keys pressed in the middle frame $X_t$. Mathematically we estimate the conditional probability of which keys are pressed at frame $t$ given video frames $X_{t-2:t+2}$. The probability of estimated keys at frame $t$ will be $P(\hat{M}_{:,t}) = P(M_{:,t} | X_{t-2,t-1,t,t+1,t+2})$. We find that the use of consecutive frames is crucial to detect changes in the pressed keys. Note that estimating all pressed keys at each frame is a harder task compared to \cite{koepke2020sight} which predicts onset events only (which and when a key is being pressed). We use ResNet18 as the backbone similar to \cite{koepke2020sight} but our architecture takes into consideration the natural phenomena appearing in this task: $1$) the visual cues of sustained keys are relatively small compared to other objects in the image such as hands and fingers; $2$) at each frame, the pressed keys may have correlation due to the concept of musical harmony so some combinations have higher chance to appear at the same time than others; $3$) the spatial dependencies are significant to detect the sustained keys but typical CNN are designed to be invariant to spatial positioning. To address these issues, we design a multi-scale feature attention network similar to \cite{yan2019multi}. Specifically, using ResNet18 as backbone, Video2Roll Net contains three functional modules: feature transform, feature refinement and correlation learning. The feature refinement setup is similar to a feature pyramid network (FPN) \cite{lin2017feature} which uses top-down features propagation mechanism. The main difference to common FPN is that in our Video2Roll Net multi-scale features at residual blocks are first transformed and re-calibrated via feature transform module before passing to the next stage. This allows the network to detect the visual cues on various scales much better. As a final component, the correlation learning module is used to learn feature spatial dependencies and semantic relevance by self-attention mechanism. Since detection of pressed keys is crucial to generate meaningful music, using multi-scale feature attention strategy enables Video2Roll Net to find the region of visual cues of key-pressed more accurately (as shown in Fig. \ref{fig:feature_map}).

\begin{figure}
    \centering
    \includegraphics[width=0.8\linewidth]{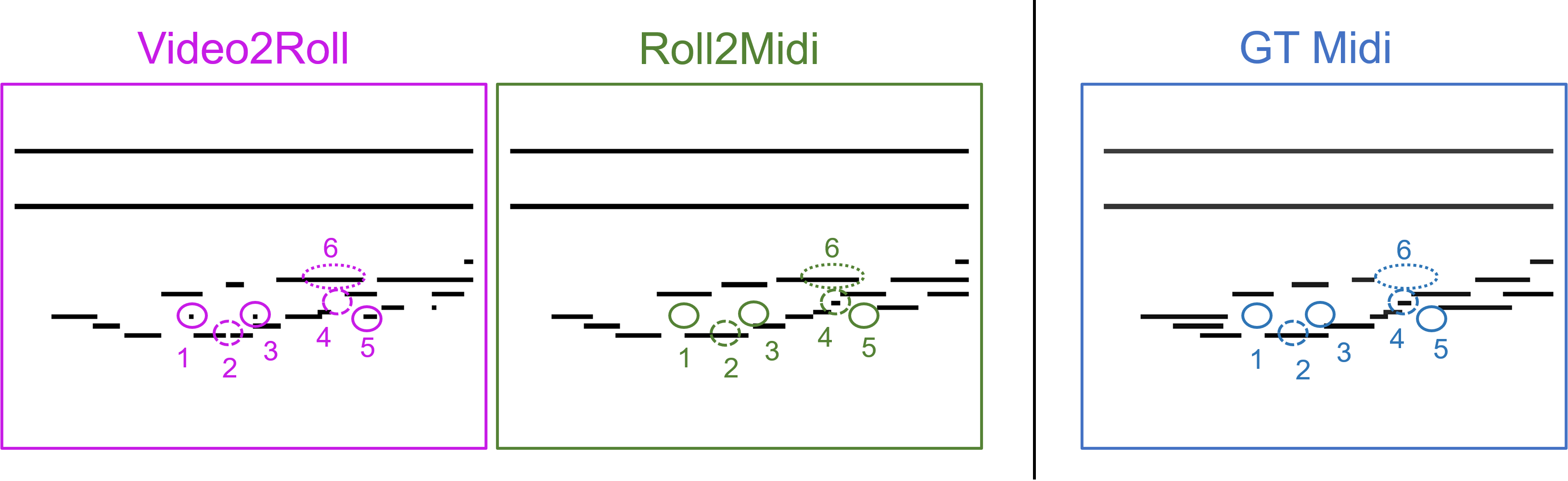}
    \caption{Comparison of Roll, Midi and Pseudo GT Midi. Solid ellipses (1,3,5) : elimination of false positives; Dashed ellipses (2,4): elimination of false negatives; Dotted ellipse (6): failure not eliminated.}
    \label{fig:MIDI_compare}
\end{figure}

\begin{figure}
    \centering
    \includegraphics[width=0.8\linewidth]{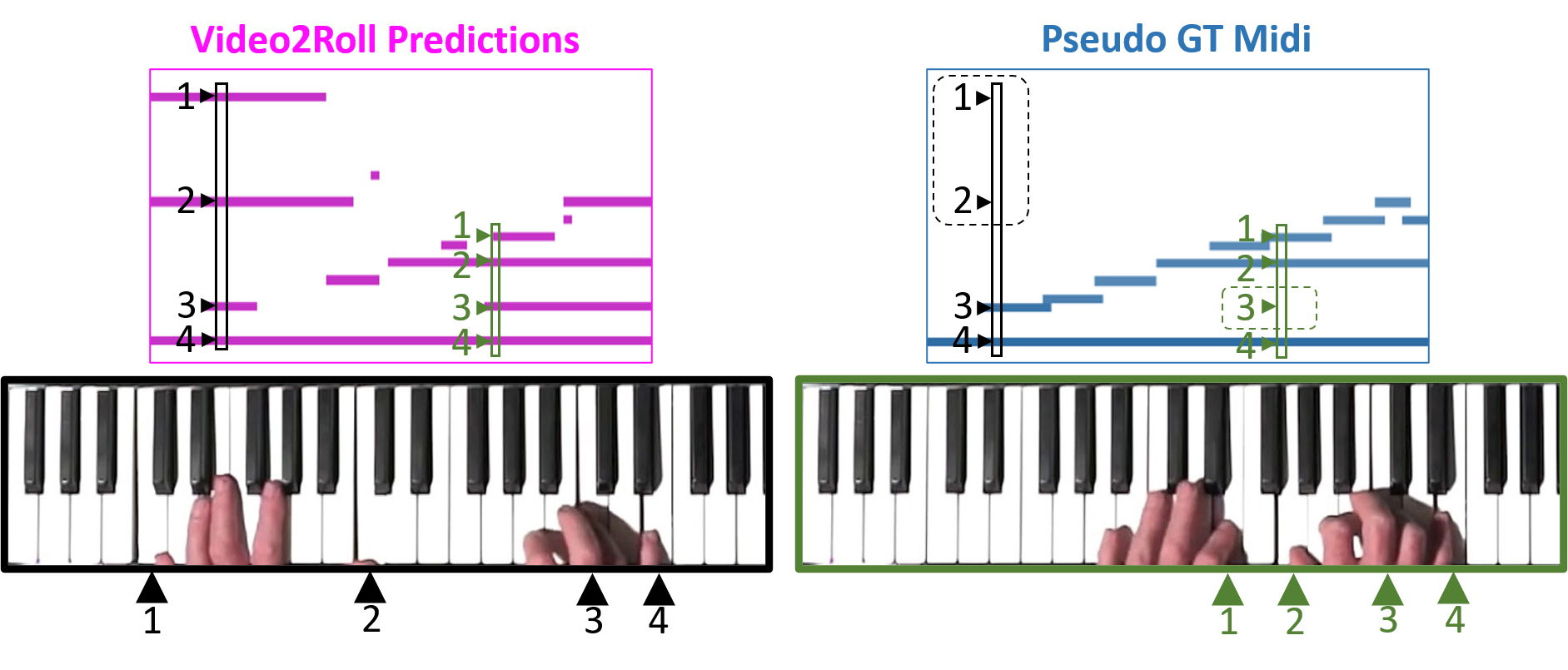}
    \caption{Examples of Pseudo GT Midi mismatches with pressed keys. Keys that are active in our predictions and in video frames (black and green) are marked as off in Pseudo GT Midi (dashed). Keyboard images are from \href{https://www.youtube.com/user/PaulBartonPiano}{Paul Barton YouTube Channel} with permission.}
    \label{fig:adaptedMIDI}
\end{figure}

\textbf{Roll2Midi Net:} The Roll prediction $\hat{M}_{:,1:T}$ of Video2Roll Net is not perfect due to various challenges. For example, hand occlusions in video frames pertain Video2Roll Net from detecting changes in pressed keys. Moreover, because $\hat{M}_{:,t}$ is predicted at each frame individually, Roll predictions do not have temporal correlation. In addition, since pseudo GT Midi is generated from Onset and Frames framework \cite{hawthorne2017onsets} which depends on the audio stream, one common phenomenon that appears is: if the performer sustains a key for sufficiently long time, the magnitude of the corresponding frequency will gradually decay to zero and this key in pseudo GT Midi will be marked as off afterwards, however, since our Video2Roll Net depends on visual information only, all pressed keys are still considered as active but this prediction will not match the reality of the audio. Examples can be seen in Fig. \ref{fig:adaptedMIDI} where in both black and green frames, Video2Roll Net detects more active keys than in Pseudo GT Midi however this is since in the frames, those keys are indeed pressed. We call this fact as mismatch of audio-visual information. To mitigate these issues, we introduce a generative adversarial network (GAN) \cite{goodfellow2014generative} to refine and complete the Video2Roll results $\hat{M}_{:,1:T}$ so that the outputs are closer to pseudo GT Midi. The GAN includes a generator $G$ and a discriminator $D$. The input of the generator is Roll predictions $\hat{M}_{:,T_1:T_2}$ and each column of $\hat{M}$ is the probability score retrieved from the last fully connected layer of Video2Roll Net after applying a sigmoid function. Using the probability scores instead of threshold outputs enables the generator to re-calibrate the probabilities and generate more robust Midi representation. The GAN objective is defined by:
\begin{equation}
    \min_G \max_D \mathbb{E}_{M \sim \mathcal{M}}[\log D(M)] + \mathbb{E}_{\hat{M} \sim \hat{\mathcal{M}}}[log(1-D(G(\hat{M})))].
\end{equation}
Our generator is a standard five depths U-Net \cite{ronneberger2015u} and the discriminator consists of with $5$ layers CNN. We use Mean Square Error (MSE) to optimize both the generator and the discriminator. During inference, we forward the Roll representation to the generator and obtain the refined representation (Midi) $\hat{M}_R = G(\hat{M})$. The Roll2Midi Net can boost the correctness of overall predictions and the estimated Midi is sufficient to be synthesized to get meaningful music close to the ground truth. Fig. \ref{fig:MIDI_compare} shows that Roll2Midi can partially eliminate false positive and false negative in Roll.
\begin{figure}
    \centering
    \includegraphics[width=0.8\linewidth]{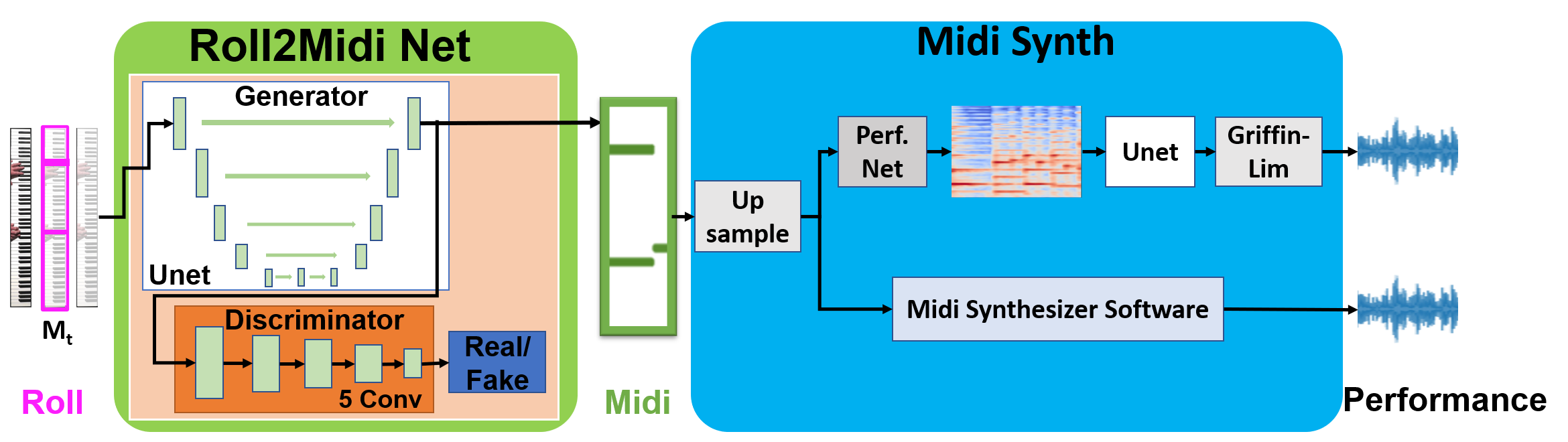}
    \caption{Detail schematic of Roll2Midi Net and Midi Synth components of \textit{Audeo} system.}
    \label{fig:ROLL2MIDI}
\end{figure}

\textbf{Midi Synth:} Both Roll and Midi can be synthesized to audio using classical Midi synthesizer. We find that it is sufficient to get clear, robust and reasonable music with predicted Midi. Moreover, classical Midi synthesizer is flexible and can support creative applications. For example, music with various timbre can be generated using piano performance video only by simply setting instruments other than piano during the synthesis step. While interesting results can be obtained at this point, the audio synthesized from classical Midi synthesizer is mechanical. To step further, we investigate whether we can generate more realistic music with Midi predictions via deep synthesizer. To do that, we pretrain a PerfNet \cite{wang2019performancenet} with Pseudo GT Midi $M$. The PerfNet learns a transformation $H$ between $M$ and spectrogram $S$. With the pretrained PerfNet, we forward propagate the Midi $\hat{M}_R$ to obtain initial estimated spectrogram $\hat{S}_R = H(\hat{M}_R)$. Note that even though our predicted Midi has been refined, discrepancy between $M$ and $\hat{M}_R$ would still exist and we find that using PerfNet to learn transformation from $\hat{M}_R$ to $S$ directly can't be generalized. We conjecture that this is due to the sensitivity of the transformation between Midi and spectrogram which increases the difficulty in the generalization. To mitigate this problem, we train one more standard U-Net to do the refinement on spectrogram level. This U-Net can be formulated as a function $U$ and we aim to minimize the L$1$ distance between $\hat{S}_R$ and $S$:
$ L_1 (\hat{S}_R,S) =  \| U(\hat{S}_R)-S  \|.$ We find that estimating initial rough spectrogram first and then perform the refinement on the spectrogram level later allows for generalization. As a last step, Griffin-Lim algorithm is used to convert the spectrogram to audio waveform \cite{griffin1984signal}.
\section{Experiments \& Results}
\textbf{Datasets.} We collect piano performance videos directly from YouTube to evaluate \textit{Audeo} pipeline. The minor constraint for data collection is top-view piano performance with fully visible keyboard. We use videos recorded by Paul Barton\footnote{\url{https://www.youtube.com/user/PaulBartonPiano}} at the frame rate $25$ fps and audio sampling rate at $16$kHz. The pseudo GT Midis are obtained via Onsets and Frames framework \cite{hawthorne2017onsets}. All Pseudo GT Midis are set binary and down-sampled to $25$ fps. We crop all videos and keep full keyboard only and remove all frames that do not contribute to piano performance (e.g. logos, black screen). We trim the initial silent sections up to the first key-pressed frame to align the video, pseudo GT Midi and audio but all silent frames within each performance are kept. Two evaluation sets are used in the experiment.\\
\textbf{\textit{Midi Evaluation Set:}} This set is to evaluate our predictions in Video2Roll Net and Roll2Midi Net. We use $24$ videos of Bach The Well-Tempered Clavier Book One (WTC B$1$) including $115$ minutes in total for training. The testing set contains first $3$ Prelude and Fugue performances of Bach The Well-Tempered Clavier Book Two (WTC B$2$) including $12$ and a half minutes in total. This results in $172,404$ training images and $18,788$ testing images.\\
\textbf{\textit{Audio Evaluation Set:}} This set is for audio evaluation only. Our aim is to test whether the generated music can be detected by music identification software. This test set contains $35$ videos from WTC B$2$ ($24$ Prelude and Fugue pairs and their $11$ variants), $8$ videos from WTC B$1$ variants and $9$ videos from other composers. This combination results in $52$ videos and $297$ minutes in total.

\textbf{Evaluation Metrics.} For \textbf{\textit{Midi Evaluation}}, we evaluate predictions from our Video2Roll Net and Roll2Midi Net by reporting precision, recall, accuracy and F1 score on frame-level defined in  \cite{bay2009evaluation}. To compare with other methods, we reproduce proposed models in  \cite{koepke2020sight} and test them on our Midi Evaluation set. For \textbf{\textit{Audio Evaluation}}, We use popular music identification App SoundHound\footnote{\url{https://www.soundhound.com/}} to perform detection test on generated music. We split every performance into multiple $20$ seconds segments and perform the detection test on every segment once. The detection is marked as success if SoundHound successfully shows correct source name of the music and failure if nothing or wrong source shows up. We report the average detected rate at segment level.
\subsection{Implementation details}
\textbf{Video2Roll Net:} All images are set in gray scale and resized into $100\times 900$. Due to the extremely imbalanced label classes in Midi evaluation set, we force each training mini-batch to contain classes evenly by over/down sampling strategy. Features obtained at residual blocks are used to do feature level transform and refinement except the first block. We train the network using binary cross entropy loss with batch size $64$.\\
\textbf{Roll2Midi Net:} We extract probability scores (without threshold-ed) from Video2Roll Net at each frame and concatenate them as Roll representation. We use a $4$ seconds Roll ($100$ frames) and slide with two seconds window during training. The five depth U-net generator takes one channel input and each depth down-samples the height and width by half. The discriminator includes five convolution layers that take the Midi as input and classify it as real or fake. Both the generator and the discriminator are trained with MSE loss with batch size $64$.\\
\textbf{Midi Synth:} We use FluidSynth\cite{henningsson2011fluidsynth} as classical Midi synthesizer. For all results, we set initial tempo to be $80$ and velocity for all active keys be $100$. For the deep synthesizer, a PerfNet is pretrained with pseudo GT Midi using MSE loss with batch size $16$. The target spectrogram of an audio clip is the magnitude part of its short-time Fourier Transform. We compute the log-scaled spectrogram with $2,048$ window size and $256$ hop size, leading to a $1025 \times 126$ spectrogram for $2$ seconds audio sampled at $16$kHz. The $2$ seconds Midi ($50$ frames) will be firstly up-sampled to $126$ frames to fit with the input shape size of PerfNet. Once we obtain the initial spectrograms from PerfNet, we train a standard five depths Unet to refine the spectrogram. Since highest frequency on a piano key is $4186.01$ Hz, we use only the frequency bins up to $576$ for training. In the end, we use Griffin-Lim algorithm\cite{griffin1984signal} to generate the final audio.\\
All networks in \textit{Audeo} system are trained in PyTorch \cite{paszke2017automatic} using Adam optimizer \cite{kingma2014adam} with $\beta_1 = 0.9$, $\beta_2 = 0.999$. For all models we use learning rate starting from $0.001$ and gradually decreasing if validation loss is stuck in plateau. Two Nvidia Titan X are used to train all components in \textit{Audeo} system. More specific implementation details can be found in Supplementary material. All code will be openly available in the future.
\subsection{Results}
% \begin{figure}
%     \centering
%     \includegraphics[width=0.8\linewidth]{new_PseudoGTMIDI_mismatch.png}
%     \caption{Examples of Pseudo GT Midi mismatches with pressed keys. Keys that are active in our predictions and in video frames (black and green) are marked as off in Pseudo GT Midi (dashed).}
%     \label{fig:adaptedMIDI}
% \end{figure}

% \usepackage{graphicx}
\begin{table}[]
\small
\centering
% \resizebox{\textwidth}{!}{%
\begin{tabular}{|l|l|l|l|l|}
\hline
Model                      & Precision     & Recall        & Accuracy & F1-score      \\ \hline
ResNet \cite{koepke2020sight} & 64.3          & 54.7        &40.4      &  49.7          \\ \hline
ResNet+Aggregation+slope \cite{koepke2020sight} & 61.5       & 57.3     &  41.2      & 50.8 \\ \hline
Video2Roll Net (Our)         & 61.2          & 65.6          &  46.4    &  56.4          \\ \hline
Roll2Midi Net TS=0.4 (Our) & 60.0 & \textbf{77.0} &    \textbf{50.6}      & \textbf{61.5} \\ \hline
Roll2Midi Net TS=0.5 (Our) & \textbf{65.1} & \textbf{69.9} &    \textbf{50.8}      & \textbf{60.4} \\ \hline
\end{tabular}%
% }
\caption{Precision, recall, accuracy and F1-score in (\%) for Midi evaluation. If not specified, all results use threshold (TS) = $0.4$ after the application of the sigmoid function.}
\label{tab:1}
\end{table}
\textbf{Midi Evaluation:} Table \ref{tab:1} shows the results of \textit{Audeo} in generation of Roll and Midi compared to other methods. The Video2Roll Net detects detailed visual cues which result in higher recall, accuracy and F1-score compared to previous works. It turns out that having less false negatives is crucial to generate complete melody without missing the notes. In addition, the relatively low precision still reflects the benefit of Video2Roll Net because mismatches in audio-visual information are a common phenomenon (See Fig. \ref{fig:adaptedMIDI}). Thereby, many false positives in our predictions are not visually wrong. Notably, music generation from visual information is nontrivial and this is one of the common challenges therefore we are not surprised by the performance but on the other hand believe that it will be enhanced in the future. To get cleaner and robust symbolic representation, Roll2Midi Net is necessary. The core of generative adversarial network enables Roll2Midi Net to partially eliminate both false negative and false positive by judging whether the generated Midi is real enough. Indeed, Roll2Midi Net boosts the overall performance even further. The F1 score of Roll2Midi outperforms the best model in \cite{koepke2020sight} by more than $10\%$.

\textbf{Audio Evaluation on Music Identification:}
\begin{table}[]
\small
\centering
\begin{tabular}{|l|c|c|c|}
\hline
 & \multicolumn{1}{l|}{Total} & \multicolumn{1}{l|}{Bach WTC B1 Variants} & \multicolumn{1}{l|}{Bach WTC B2 \& Other } \\ \hline
ResNet+FluidSynth & 55.9        &  74.2          &  52.9      \\ \hline
Roll+FluidSynth   & 62.6         & 79.6          &  59.6      \\ \hline
Midi+PerfNet      & 73.0          & 80.6         &  71.6          \\ \hline
Midi+FluidSynth   &  \textbf{73.9} & \textbf{85.6} & \textbf{72.4} \\ \hline\hline
Ground Truth      & 89.2         & 92.6           & 87.7           \\ \hline
\end{tabular}
\caption{Sound Hound music identification rate in (\%).}
\label{tab:SH Test}
\end{table}
\begin{figure}[t]
    \centering
    \includegraphics[width=0.7\linewidth]{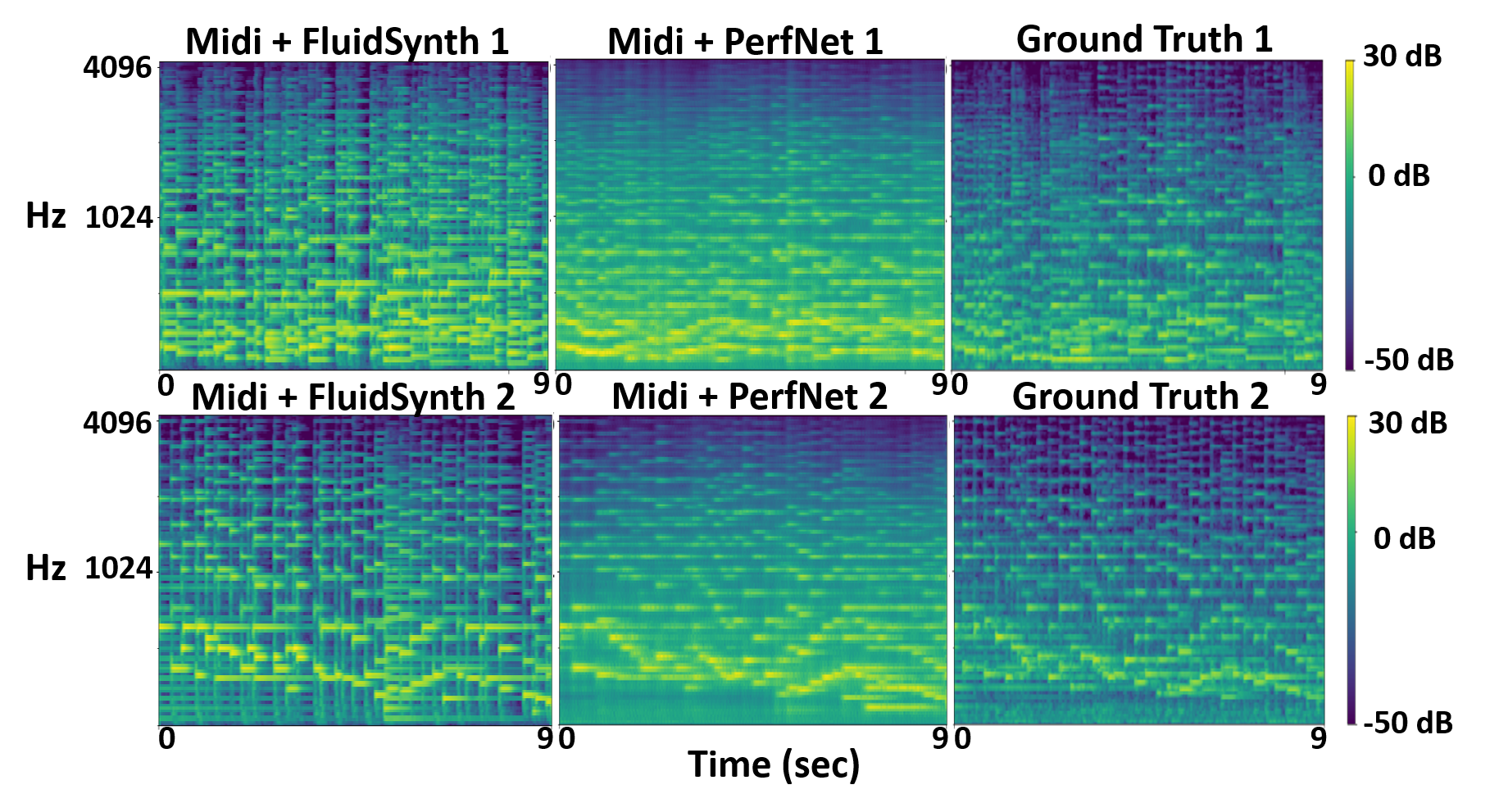}
    \caption{Two samples of generated spectrograms vs Ground Truth.}
    \label{fig:spec}
\end{figure}
We compare the detection by SoundHound of samples generated from \textit{Audeo} system to ResNet baseline and to ground truth audio. Furthermore, we synthesize Roll and Midi obtained from \textit{Audeo} via FluidSynth or PerfNet to test and exclude synthesizer effects. The results of music identification are shown in Table \ref{tab:SH Test}. Note that the Bach WTC B$1$ have already been learned during training and we use their variants to evaluate whether \textit{Audeo} is robust to different performance styles such as fast tempo, staccato, legato and so on. It turns out that all \textit{Audeo} methods outperform the ResNet baseline and synthesizing Midi via FluidSynth or PerfNet can reach more than $80\%$ detected rate while Midi+FluidSynth achieves the best accuracy ($85.6\%$). This is compared the the Ground Truth detection of ($92.6\%$). This indicates that \textit{Audeo} is able to capture the core of learned music and is not sensitive to variance in performance. For test videos from the type that was not introduced in training at all, such as Scott Joplin, both Midi+PerfNet and Midi+FluidSynth pass $70\%$ detected rate while ResNet baseline obtains $52.9\%$. While the gap with the ground truth ($72.4$ vs $87.7\%$) is still obvious, the identification results demonstrate the robustness and generality of the \textit{Audeo} system. In terms of total average, Midi+FluidSynth performs better than other methods and outperforms ResNet baseline by $18\%$. Notably, using PerfNet as synthesizer results with slightly lower detection than FluidSynth in this test. While deep synthesizer may recover emotion and naturalness in the spectrogram domain, it also introduces noise which is non-trivial to reduce. Fig.\ref{fig:spec} compares spectrograms of samples synthesized via FluidSynth and PerfNet. While both syntheses produce similar spectrograms to the ground truth one, we observe that Midi+FluidSynth is cleaner but having same notes velocity everywhere results in unnatural sound. On the other hand, Midi+PerfNet has magnitude variance and changes smoothly in time but noticeable noise exists.

\section{Conclusion}
We present a novel one of a kind full pipeline system \textit{Audeo} for generating music from silent piano performance video. Each component in \textit{Audeo} is interpretable and flexible to be used for various practical purposes such as key detection, piano learning synchronization, timbre modulation, etc. Experimental results demonstrate that \textit{Audeo} can effectively generate reasonable music that can be detected by music identification software.

% \section*{Broad Impact}
% The \textit{Audeo} system enables music generation from silent piano performance video. One classical application is to recover a corrupted audio channel in a piano performance video. Moreover, since \textit{Audeo} uses Midi as an intermediate representation, this provides a large amount of possibilities to manipulate the generated Midi creatively. For example, people could use the predicted Midi to synthesize music of any instrument by just giving a piano performance video. This can be also extended to virtual piano environment in the real-time where \textit{Audeo} can generate music from visual information when there is no real sound available at all. All these directions would benefit from \textit{Audeo}. Due to the fact that the generated music can be detected by a music identification App, one concern could be the possibility that a fake pianist could scam audiences utilizing \textit{Audeo} system. This is a common concern in the application of any generative model. Failure in \textit{Audeo} may bring up unsatisfying music but we do not expect serious consequences. Also, while \textit{Audeo} is trained and tested on videos of the same pianist, we believe the full pipeline is valid and robust in general due to the variance in the pianist can only result in the difference in hand shape and this variance can easily be incorporated in \textit{Audeo} by either fine tuning or adding more data to the training set.

\bibliographystyle{unsrt}
\end{document}